\pdfoutput=1

\documentclass[oneside,reqno]{amsart}

\usepackage{amsaddr}
\usepackage{flushend}
\usepackage{lipsum} 
\usepackage{amsmath}
\usepackage{subfig}
\usepackage{tabularx}
\usepackage{booktabs}
\usepackage{multicol}
\usepackage{pgfplots}
\usetikzlibrary{positioning,shadows,patterns,pgfplots.groupplots}
\usepackage{cleveref}
\usepackage{epstopdf}
\usepackage{algorithm}
\usepackage{algpseudocode}
\usepackage{caption}
\usepackage{longtable}
\usepackage{float}
\usepackage{xcolor}

\usepackage{listofitems,readarray,environ,filecontents,tabstackengine,etoolbox}
\usetikzlibrary{snakes}

\usepackage[
hmarginratio={1:1},     
vmarginratio={1:1},     
heightrounded,          
]{geometry}

\definecolor{bblue}{HTML}{4F81BD}
\definecolor{rred}{HTML}{C0504D}
\definecolor{ggreen}{HTML}{9BBB59}
\definecolor{ppurple}{HTML}{9F4C7C}

\Crefname{equation}{Eq.}{Eqs.}
\Crefname{figure}{Figure}{Figures}
\Crefname{tabular}{Table}{Tables}

\usepackage[backend=bibtex,
	style=nature,
	sorting=none,
	giveninits=true,
	isbn=false,doi=false,url=true,
	natbib=true
]{biblatex}

\DeclareFieldFormat{journaltitle}{#1\isdot}
\DeclareFieldFormat[article,periodical]{volume}{#1\isdot}

\begin{document}
	\title[Machine Assistance for Credit Card Approval]{Machine Assistance for Credit Card Approval? Random Wheel can	Recommend and Explain}
	
	\author{Anupam Khan *}
	\address{Department of Computer Science and Engineering, Indian Institute of Technology Kharagpur, West Bengal 721302, India}
	\email{anupamkh@iitkgp.ac.in}
	\thanks{* Corresponding author}
	
	\author{Soumya K. Ghosh}
	\address{Department of Computer Science and Engineering, Indian Institute of Technology Kharagpur, West Bengal 721302, India}
	\email{skg@cse.iitkgp.ac.in}

	\begin{abstract}
		Approval of credit card application is one of the censorious business decision the bankers are usually taking regularly. The growing number of new card applications and the enormous outstanding amount of credit card bills during the recent pandemic make this even more challenging nowadays. Some of the previous studies suggest the usage of machine intelligence for automating the approval process to mitigate this challenge. However, the effectiveness of such automation may depend on the richness of the training dataset and model efficiency. We have recently developed a novel classifier named \textit{random wheel} which provides a more interpretable output. In this work, we have used an enhanced version of random wheel to facilitate a trustworthy recommendation for credit card approval process. It not only produces more accurate and precise recommendation but also provides an interpretable confidence measure. Besides, it explains the machine recommendation for each credit card application as well. The availability of recommendation confidence and explanation could bring more trust in the machine provided intelligence which in turn can enhance the efficiency of the credit card approval process.
	\end{abstract}

	\keywords{Credit card approval, predictive modelling, explainable AI, classification}

	\maketitle

\section{Introduction}
In August 2019, The Indian Express reported nearly 24.4\% and 71\% increase in credit card outstanding amount in India as compared to 2018 and 2017 respectively \cite{indexpress2019credit}. The Nilson Report \cite{nelson2021credit} recently found a similar observation in the United States where 24\% of total credit card purchase in 2020 were outstanding. Although the percentage declined from 28\%, as observed in 2019, it amounts to 866 billion dollars in total. Surprisingly, this value is more than the GDP of many countries like Saudi Arabia, Turkey, Switzerland, Poland, Thailand, Sweden, Belgium etc. \cite{worldbank2021gdp}. Credit cards obtained through fraudulent applications may contribute a considerable share to this outstanding amount. Approval of credit card is, in fact, one of the critical business decision the bankers are taking everyday. Although their experience makes them proficient in this job, it is not possible always to take correct decisions. The magnitude of correctness could decrease with the increasing volume of applications to be processed daily. Interestingly, the recent pandemic has further increased the volume of digital transactions and it could magnify the demand of new credit card applications to a considerable extent \cite{creditcardsurge2020pandemic}. The use of machine intelligence may help in this regard.

\begin{figure}[t]
	\centering
	\includegraphics[width=\textwidth]{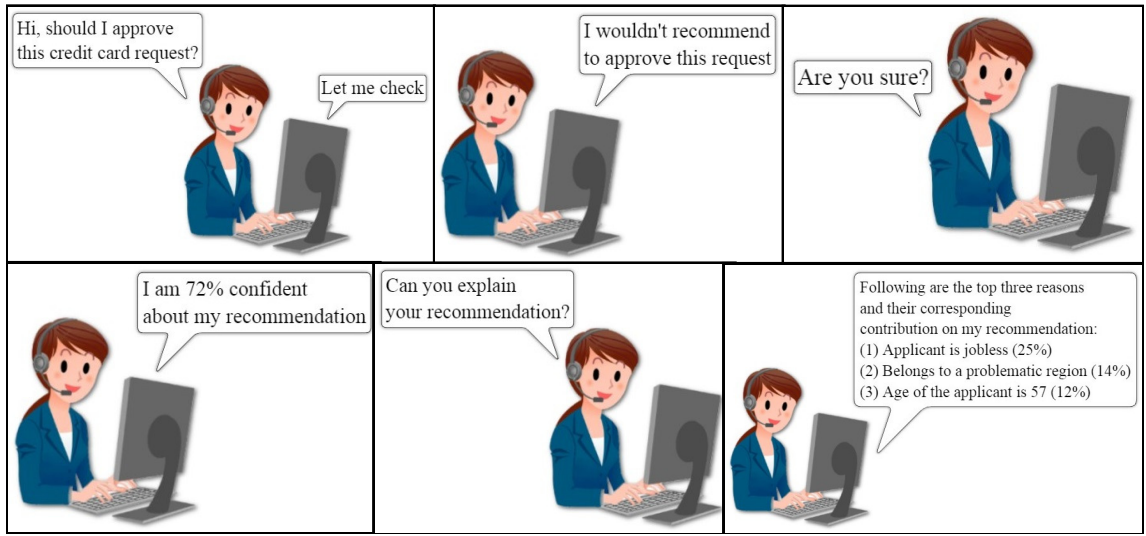}
	\caption{Overview of the proposed intelligent recommendation system}
	\label{fig:motivation-story}
\end{figure}

Several studies \cite{zarnaz2021credit, duan2020performance, sakprasat2007classification} in the literature suggest using machine intelligence for automating the credit card approval process. However, such automation may not be effective enough in a real banking environment. It is important to mention here that machines are not born intelligent \cite{warwick2012not}. The supervised learning algorithms generally train them to be intelligent using the knowledge extracted from historical data. The machines are therefore highly probable to be biased with the historical data and learning algorithms \cite{mehrabi2019survey}. The bias could make a machine incapable of handling adverse situations not trained earlier. In contrast, a human being can manage such a situation by either using his/her skill or collaborating with others. So, should we use machine intelligence for the automated approval of credit cards? We believe that, instead of complete automation, machine intelligence can be used to assist human being in the credit card approval process.

Existing machine learning approaches generally assist the decision-making process by predicting or recommending the output of an observation. However, it is quite often reported in literature \cite{floridi2019establishing, toreini2020relationship, chatila2021trustworthy, jimenezluna2020drug} that the end-users are sceptical about the trustworthiness of such recommendation. It may be more prevalent in sensitive areas like finance, healthcare etc. Significantly, it is not possible for a machine to correctly recommend the approval of all credit card applications. Even if a machine is tested to be sufficiently accurate, unexpected behaviour could be possible in a real banking environment. The availability of recommendation confidence can help in such circumstances. Low confidence can alert the end-user about possible false recommendation. Another potential reason behind the lack of trustworthiness is the black-box nature of intelligent algorithms available nowadays \cite{setzu2021glocalx, adadi2018peeking, lundberg2020explainable}. Researchers have used various state-of-the-art classification approaches like decision tree \cite{kumari2019analysis}, support vector machine \cite{plawiak2019application, rodan2016credit}, artificial neural network \cite{ilgun2014application, khan2016chicken}, genetic programming \cite{sakprasat2007classification, le2016improving, srinivasan2018multi}, deep learning \cite{markova2021credit, kibria2021application, neagoe2018deep} etc. for predicting credit card approvals. In most cases, these machines fail to explain why a certain credit card application is recommended to be approved or rejected. At least we could not find any study which explains the machine recommendation in literature.

Here, we present a completely different perspective of machine intelligence for credit card approval. Achieving an acceptable recommendation accuracy is not the only objective of this study. It rather aims to present a more intelligent and trustworthy recommendation approach. \Cref{fig:motivation-story} graphically outlines the broad objective of the proposed intelligent recommender system for this purpose. Although various state-of-the-art classifiers could achieve significant accuracy, extracting recommendation confidence and explanation from these classifiers is a research challenge. We have recently proposed a novel classifier, \textit{random wheel} \cite{khan2021randomwheel}, that not only provides such confidence but also has the potential of explaining the classification. However, it works with categorical data only. The dataset used here for validating our methods contains both categorical and numerical attributes. In this work, we have therefore enhanced the random wheel to make it compatible with the mixed dataset. Significantly, this enhancement considers input data in their actual format without converting them to either categorical or numeric form. The enhanced random wheel turned out to be better than state-of-the-art classifiers. Moreover, it additionally extracts the explanations of each individual credit card approval or rejection recommendation.

\section{Results}  \label{sec:results}
This section presents the result obtained while evaluating our methods with the Australian credit card approval dataset \cite{creditapprovalucidataset}. This widely used dataset is publicly available in the University of California Irvine (UCI) machine learning repository \cite{ucimlrepo}. As presented in \Cref{tab:dataset}, it contains fifteen attributes ($A01 - A15$) and one class variable ($A16$). The attribute names and their values were changed to meaningless symbols to protect the confidentiality of the data.  Nine of these attributes ($A01, A04, A05, A06, A07, A09, A10, A12, A13$) are categorical. Amongst the other six attributes, three ($A11, A14, A15$) contains integer values and the rest three ($A02, A03, A08$) hold real numbers. A few missing values are also observed for these attributes. The UCI repository \cite{creditapprovalucidataset} mentions this dataset as an interesting one due to the mixture of continuous and nominal attributes with both small and large numbers of values. Besides, the class variable contains two values: (i) + (positive) where credit card applications are approved, and (ii) $-$ (negative) where the applications are rejected. In this dataset, credit card applications are approved for 44.49\% cases whereas the rest 55.51\% applications are rejected.

\begin{table}
	\centering
	\captionof{table}{Summary of credit card approval dataset} 
	\label{tab:dataset}
	\scriptsize
	\begin{tabular}{ll}
		\toprule
		\bfseries{Description}		& \bfseries{Value}					\\
		\midrule
		Number of Attributes		& 15								\\
		Type of Attributes 			& Categorical, Integer, Real		\\
		Number of Instances 		& 690								\\
		Missing Values				& Present							\\
		Class labels				& 2 ($+$ / positive, $-$ / negative)	\\
		Class Distribution			& positive: 44.49\%, negative: 55.51\%	\\
		\bottomrule
	\end{tabular}
\end{table}

In this work, we compare the performance of the random wheel with other classifiers available in the literature. We have trained the model and extracted the effectiveness metrics using 10-fold cross-validation method. Importantly, the depth and noise fraction of the random wheel are defined as 3 and 0.5 in this study respectively. We have considered 100 trials for measuring factor importance and recommending an observation. The following part of this section presents the evaluation results.

\subsection{Comparison with state-of-the-art classifiers}

The random wheel is not an enhanced version of existing classifier. It rather targets to classify an unknown observation from a completely different perspective. We have therefore compared the effectiveness of random wheel for credit card approval recommendation with following ten state-of-the-art classifiers: (i) na\"ive Bayes \cite{john2013estimating}, (ii) Bayesian network \cite{heckerman1995bayesian}, (iii) logistic regression \cite{le1992ridge}, (iv) decision tree \cite{quinlan2014c4}, (v) support vector machine \cite{zeng2008fast}, (vi) k-nearest neighbour \cite{aha1991instance}, (vii) artificial neural network \cite{rosenblatt1958perceptron}, (viii) random forest \cite{breiman2001random}, (ix) deep learning \cite{lecun2015deep}, and (x) boosting method \cite{freund1996experiments}. The default Weka implementation of these classifiers are considered for this comparison. 
\par
\begin{figure}[t]
	\centering
	
		\pgfplotsset{width=\textwidth, height=4.0cm}
		\begin{tikzpicture}
			\begin{axis}[
				ybar, ymin=0.5, ymax=0.9,
				bar width=4pt,
				ymajorgrids,
				legend style={at={(0.5,1.00)},anchor=south,legend columns=-1},
				xlabel={Classifier}, x label style={at={(0.5,-0.15)}, font=\footnotesize},
				xtick=data,
				xticklabels={NB, BN, LR, DT, SVM, kNN, ANN, RF, DL, BM, RW},
				x tick label style={align=center, font=\scriptsize},
				y tick label style={align=center, font=\scriptsize},
				major x tick style = {opacity=0},
				minor x tick num = 1,
				minor tick length=2ex,
				]
				\addplot[draw=black,fill=ggreen] table[x index=0,y index=1] \effdataset; 
				\addplot[draw=black,fill=ppurple] table[x index=0,y index=2] \effdataset; 
				\addplot[draw=black,fill=bblue] table[x index=0,y index=3] \effdataset; 
				\addplot[draw=black,fill=rred] table[x index=0,y index=4] \effdataset; 
				\legend{\scriptsize Accuracy, \scriptsize Precision, \scriptsize F-measure, \scriptsize Kappa}
			\end{axis}
		\end{tikzpicture}
	
	\caption{Comparison of different classifiers}
	\label{fig:classifier-comparision}
\end{figure}
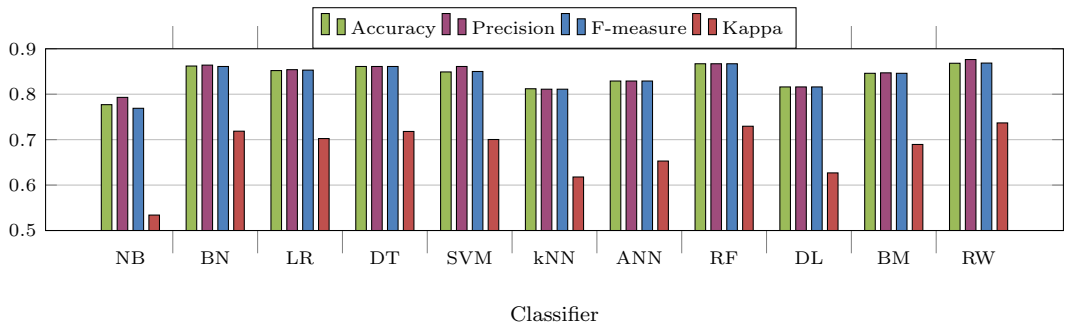

Significantly, the random wheel turns out to be marginally better than the above state-of-the-art classifiers. As presented in \Cref{tab:classifier-comparison}, the random wheel yields the highest accuracy, precision, F-measure, and kappa of 0.8681, 0.8763, 0.8685, and 0.7368 respectively. The random forest is the closed competitor in this regard. \Cref{fig:classifier-comparision} additionally compares the effectiveness of various classifiers graphically. All classifiers except na\"ive Bayes are able to achieve more than 80\% accuracy, precision, and F-measure. However, only Bayesian network, logistic regression, decision tree, support vector machine, random forest and random wheel report kappa measure of more than 0.7000.

\pgfplotstableread{
	0	0.7770	0.7930	0.7690	0.5340
	1	0.8620	0.8640	0.8610	0.7186
	2	0.8520	0.8540	0.8530	0.7024
	3	0.8610	0.8610	0.8610	0.7180
	4	0.8490	0.8610	0.8500	0.7003
	5	0.8120	0.8110	0.8110	0.6178
	6	0.8290	0.8290	0.8290	0.6529
	7	0.8670	0.8670	0.8670	0.7295
	8	0.8160	0.8160	0.8160	0.6268
	9	0.8460	0.8470	0.8460	0.6894
	10	0.8681	0.8763	0.8685	0.7368
}\effdataset

\begin{table}[t]
	\centering
	\caption{Recommendation efficiency observed in different classifiers} 
	\label{tab:classifier-comparison}
	\scriptsize
	\begin{tabular}{lllllll}
		\toprule
		\bfseries{Approach}			& \bfseries{Classifier in Weka}		& \bfseries{Code}		& \bfseries{Accuracy}	& \bfseries{Precision}	& \bfseries{F-measure}		& \bfseries{Kappa}		\\
		\midrule
		Na\"ive bayes				& NaiveBayes			& NB	& 0.7770	& 0.7930	& 0.7690	& 0.5340		\\
		Bayesian network			& BayesNet				& BN	& 0.8620	& 0.8640	& 0.8610	& 0.7186		\\
		Logistic regression			& Logistic				& LR	& 0.8520	& 0.8540	& 0.8530	& 0.7024		\\
		Decision tree				& J48					& DT	& 0.8610	& 0.8610	& 0.8610	& 0.7180		\\
		Support vector machine		& SMO					& SVM	& 0.8490	& 0.8610	& 0.8500	& 0.7003		\\
		K-nearest neighbour			& IBk					& kNN	& 0.8120	& 0.8110	& 0.8110	& 0.6178		\\
		Artificial neural network	& MultilayerPerceptron	& ANN	& 0.8290	& 0.8290	& 0.8290	& 0.6529		\\
		Random forest				& RandomForest			& RF	& 0.8670	& 0.8670	& 0.8670	& 0.7295		\\
		Deep learning				& Dl4jMlpClassifier		& DL	& 0.8160	& 0.8160	& 0.8160	& 0.6268		\\
		Boosting method				& AdaBoostM1			& BM	& 0.8460	& 0.8470	& 0.8460	& 0.6894		\\
		Random wheel				& -						& RW	& 0.8681*	& 0.8763*	& 0.8685*	& 0.7368*		\\
		\bottomrule
		* highest
	\end{tabular}
\end{table}

\subsection{Confidence}
As mentioned in \cite{khan2021randomwheel}, the random wheel provides an additional confidence measure while classifying an unknown observation. It measures the confidence by comparing the angular velocities of the winner and runner-up wheel. The confidence would be higher if the winner wheel rotates much faster than the runner-up wheel. We have compared the average confidence of correctly recommended observations and incorrect recommendations in the credit card approval dataset. The average confidence of correct recommendations turns out to be almost double concerning to the incorrectly recommended cases. An average 52.19\% confidence is observed in 599 correct recommendations. In contrast, the average confidence of 91 incorrectly recommended cases is 24.08\% only.
\par
\Cref{fig:prediction-confidence} presents a heatmap of observed confidences in decreasing order to depict the above-mentioned difference. The colour here changes from blue to yellow with decreasing value of confidence. \Cref{fig:prediction-confidence-wrong} shows the observed confidence of 91 incorrect recommendations whereas \Cref{fig:prediction-confidence-correct} presents the confidence of 599 correctly recommended observations. It is quite evident from \Cref{fig:prediction-confidence-correct} that the higher confidence is more frequent in the case of correct classifications. In contrast, a yellow colour in the major part of \Cref{fig:prediction-confidence-wrong} indicates low confidence for incorrectly classified cases.

\makeatletter\let\strippt\strip@pt\makeatother
\def\mytrunc#1.#2\relax{#1}
\def\mymult#1.#2#3#4#5\relax{%
	#1\ifx.#2000\else%
	#2\ifx.#300\else%
	#3\ifx.#40\else%
	#4\fi\fi\fi%
}

\def\cellwd{15pt}
\colorlet{plotcolmax}{cyan}
\colorlet{plotcolmin}{yellow!20}
\def\legendwd{6pt}
\def\legendht{30pt}
\newlength\dlegend
\newcounter{legcnt}
\newtoks\tabAtoks
\newcount\plotvalue
\newlength\pvlen
\newcommand\apptotoks[2]{#1\expandafter{\the#1#2}}
\NewEnviron{stackColor}[1][100]{%
	\ignoreemptyitems%
	\def\tAtmp{plotcolmax!}%
	\tabcolsep=0pt\relax%
	\setsepchar{\\/ }%
	\readlist*\tabA{\BODY}%
	\tabAtoks{}%
	\foreachitem\i\in\tabA[]{%
		\ifnum\listlen\tabA[\icnt]>1\relax%
		\foreachitem\j\in\tabA[\icnt]{%
			\expandafter\pvlen\j pt\relax%
			\edef\tmp{\mymult#1.000\relax}
			\divide\pvlen by \tmp%
			\multiply\pvlen by 100000
			\edef\tmp{\strippt\pvlen}%
			\edef\tmp{\expandafter\mytrunc\tmp.\relax}%
			\plotvalue=\tmp\relax%
			\xdef\plotmax{#1}%
			\ifnum\jcnt=1\relax\else\apptotoks\tabAtoks{&}\fi%
			\expandafter\apptotoks\expandafter\tabAtoks\expandafter{%
				\expandafter\textcolor\expandafter{\expandafter\tAtmp%
					\the\plotvalue!plotcolmin}{\rule{\cellwd}{\cellht}}}%
		}%
		\ifnum\icnt<\listlen\tabA[]\relax\apptotoks\tabAtoks{%
			\\}\fi%
		\fi%
	}%
	\def\tmp{\setstackgap{S}{0pt}\tabbedShortstack}%
	\expandafter\tmp\expandafter{\the\tabAtoks}%
}
\newcommand\plotit[2][100]{%
	\readarraysepchar{\\}%
	\readdef{#2}\mydata%
	\def\tmp{\begin{stackColor}[#1]}%
		\expandafter\tmp\mydata\end{stackColor}%
}
\newcommand\makelegend[2][\fboxrule]{{%
		\dlegend=\legendht%
		\divide\dlegend by 101%
		\setcounter{legcnt}{0}%
		\savestack\thelegend{}%
		\setstackgap{S}{0pt}%
		\whileboolexpr{test {\ifnumcomp{\thelegcnt}<{101}}}{%
			\savestack\thelegend{\stackon{\thelegend}{\textcolor{%
						plotcolmax!\thelegcnt!plotcolmin}{\rule{\legendwd}{\dlegend}}}}%
			\stepcounter{legcnt}%
		}%
		\fboxrule#1\relax\fboxsep=0pt\relax\fbox{\thelegend}%
		\def\plottick{\rule[.5\dimexpr-\dp\strutbox+\ht\strutbox]{5pt}{%
				\fboxrule}}%
		\raisebox{.5\dimexpr\dp\strutbox-\ht\strutbox-\fboxrule}{%
			\def\stackalignment{l}%
			\stackon[\dimexpr\legendht]{\smash{\plottick0}}{\smash{%
					\plottick\plotmax\ #2}}%
		}%
}}

\def\cellwd{0.6pt}
\def\cellht{40pt}
\colorlet{plotcolmax}{blue!90}
\colorlet{plotcolmin}{yellow!80}
\def\legendwd{8pt}
\def\legendht{40pt}

\begin{figure}
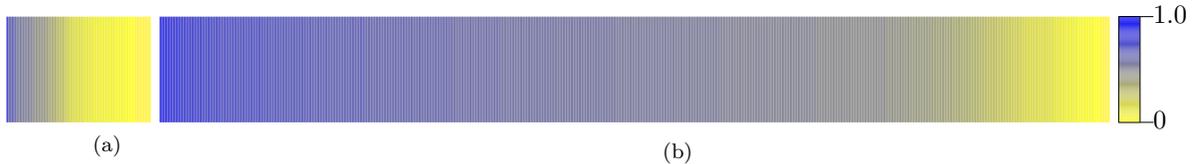
%
	\subfloat[\label{fig:prediction-confidence-wrong}]{{
		\plotit[1.0]{wrong.dat}
	}}%
	\subfloat[\label{fig:prediction-confidence-correct}]{{
		\plotit[1.0]{correct.dat}
		~\makelegend[.1pt]{}
	}}
	\caption{Observed prediction confidence for (a) incorrect recommendations, and (b) correct recommendations}%
	\label{fig:prediction-confidence}%
\end{figure}

\begin{figure}%
	\centering
	\subfloat[\centering Case-I \label{fig:sample-output-case1}]{\includegraphics[width=0.48\textwidth]{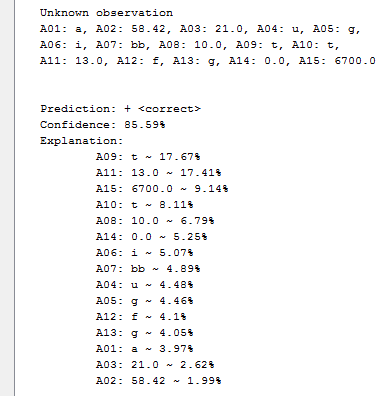} }%
	\subfloat[\centering Case-II \label{fig:sample-output-case2}]{\includegraphics[width=0.48\textwidth]{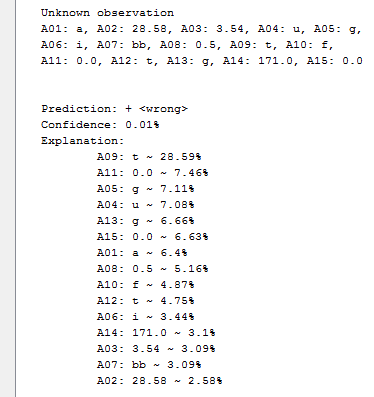} }%
	\caption{Sample output of credit card approval prediction using random wheel}%
	\label{fig:sample-output}%
\end{figure}

\subsection{Explanation}
As mentioned earlier, the objective of this work is not only to achieve better efficiency but also to provide a trustworthy explanation while recommending the credit card approvals. The wheels in a random wheel trial rotate on the application of multiple elementary forces. The randomly selected set of deciding factors determines the magnitude of these elementary forces. It is therefore quite possible to determine the contribution of each deciding factor and in turn each attribute, on the angular velocity of the wheel.

\Cref{fig:sample-output} shows the sample outputs provided by the random wheel while recommending two different observations. In Case-I (refer \Cref{fig:sample-output-case1}), it correctly recommends the credit card approval with 85.59\% confidence. The $A09$, $A11$, and $A15$ are the top three attributes that contribute 17.67\%, 17.41\%, and 9.14\% in this recommendation respectively. \Cref{fig:sample-output-case2} presents the second case where the random wheel wrongly recommends approving the credit card application. However, it is not very confident about this recommendation here. Furthermore, it recommends approving the application primarily due to $A09$ which contributes more than one-fourth of the total. The $A11$ and $A05$ are the other two most important factors here but they together contribute less than 15\% of the total.

\section{Discussion and conclusion}  \label{sec:discussion}
In this work, we have presented the effectiveness of the random wheel classifier in recommending credit card approvals. The results show that it is capable of providing more trustworthy recommendation concerning the other state-of-the-art classifiers available nowadays. The availability of additional confidence measure and explanation in the random wheel could build trust among bankers and help them in accepting the machine intelligence for credit card approvals. 
\par
Let us discuss this in light of the objectives depicted in \Cref{fig:motivation-story}. The banker may ask a machine to tell whether s/he should approve the credit card application. An intelligent machine should be capable of providing a statement like \textit{``I wouldn't recommend to approve this request''}. Importantly, a random wheel empowered machine can provide a more accurate and precise recommendation as it shows better effectiveness than other state-of-the-art classifiers. The banker may further ask whether the machine is sure about this recommendation. Most of the traditional state-of-the-art classifiers would fail to answer this. Some advanced classifiers can provide a weightage for each possible output. For example, tree augmented na\"ive Bayes \cite{jiang2012tan} provides separate weightage for each class labels during recommendation. However, multiple weights may lead to more confusion for the end-user. In contrast, the random wheel empowered machine is capable of answering like \textit{``I am 72\% confident about my recommendation''}. Such single confidence measure could be more interpretable by the banker while approving a credit card applications.
\par
Besides, the random wheel is also capable of explaining individual recommendation. The traditional machine learning algorithms generally explains the effectiveness of individual attributes by measuring the factor importance. It indicates the global explanation of how various attributes could influence the machine recommendation on a particular dataset. The literature has a rich history of such global explanations. \citet{lundberg2020explainable} rightly mentioned that we should now focus on the local explanations of input features on the individual recommendation. Significantly, a random wheel empowered machine is capable of explaining individual recommendation by extracting the contribution of each attribute on a classification task. It, therefore, provides a trustworthy explanation, for example, like \textit{``Following are the top three reasons and their corresponding contributions on my recommendation: (1) applicant is jobless (25\%), (2) belongs to a problematic region (14\%), (3) age of the applicant is 57 (12\%)''}.
\par
Availability of explanation with machine assisted recommendation could bring a radical change in the area of artificial intelligence. It could remove the reluctance on accepting the machine intelligence and help in growing trust among the beneficiaries. This has brought extensive focus on the emerging area of explainable artificial intelligence nowadays. In this work, we have presented an approach to deal with such an explanation in the banking sector. Such explanation can further help in other sensitive areas like healthcare where trust in the machine assistance is very crucial. Besides, further impact analysis of explanation availability on the banker's decision could be an interesting future research direction.

\section*{Enhanced random wheel}  \label{sec:enh-rw}
The novel wheel based classification approach \cite{khan2021randomwheel} was earlier proposed by us for predicting student performance. The concept behind this approach emerged from the rotation of multiple wheels in a bicycle race where each wheel represents a class label. These wheels rotate on the application of different resultant forces in multiple trials. Significantly, the random wheel classifier determines the magnitude of the resultant forces in a trial using a fixed set of deciding factors. These factors are the combination of different attributes available in the dataset. In $i^{th}$ trial, this novel classifier first chooses the $n_i$ (a random integer) most important deciding factors. These $n_i$ factors generate different weighted elementary forces on the wheels which causes them to rotate with different angular velocities. The classifier aggregates all angular velocities in multiple trials to finally classify an unknown observation.
\par
It is important to mention here that the earlier version of the novel random wheel classifier considered categorical attributes only. It used mean decrease Gini coefficient, scale Gini coefficient and lift of association for measuring factor importance, weightage and elementary force respectively. However, these metrics are challenging to measure in presence of continuous attributes. We have therefore adopted a different methodology to measure these metrics as the credit card approval dataset contains both categorical and continuous attributes. This enhanced version of the random wheel handles the mixed type of attributes without conversion. Here we present the factor importance and resultant force measurement methodologies in the enhanced random wheel classifier. We additionally propose the methodology to extract the explanation of machine recommendation in the latter part of this section.

\subsection*{Factor importance}
In this work, we have considered the mean decrease in the number of bins to measure the factor importance. Let us explain the concept of bin first. Say the training dataset of credit card approval contains ten records and the labels of class variable are ordered as \{$+, +, -, -, -, +, -, +, -, -$\}. A bin is a subset of ordered class labels where all items of a bin are the same. The bin changes whenever the class label switches from $+$ to $-$ or vice-versa. The above-mentioned training dataset contains six bins as the label changes on five occasions. Here, the bins are: \{\{$+, +$\}, \{$-, -, -$\}, \{$+$\}, \{$-$\}, \{$+$\}, \{$-, -$\}\}.
\par
Important factors generally take a leading role in deciding the classification output. If plotted graphically, they are more capable of grouping the identical class labels than that of a comparatively less important factor. It is therefore quite obvious that the training dataset would contain fewer bins if sorted as per the attribute values of a more important factor. The number of bins, in contrast, can increase as well when sorted as per a noisy factor. Significantly, an ideal factor should produce only $m$ bins in $m$-ary classification. It is essential to mention here that some datasets could be imbalanced in nature. These datasets themselves contain fewer bins by default. We have therefore considered both expected bin count in the training dataset and sorted dataset for importance measurement.

\begin{figure}[t]
	\centering
	\includegraphics[width=\textwidth]{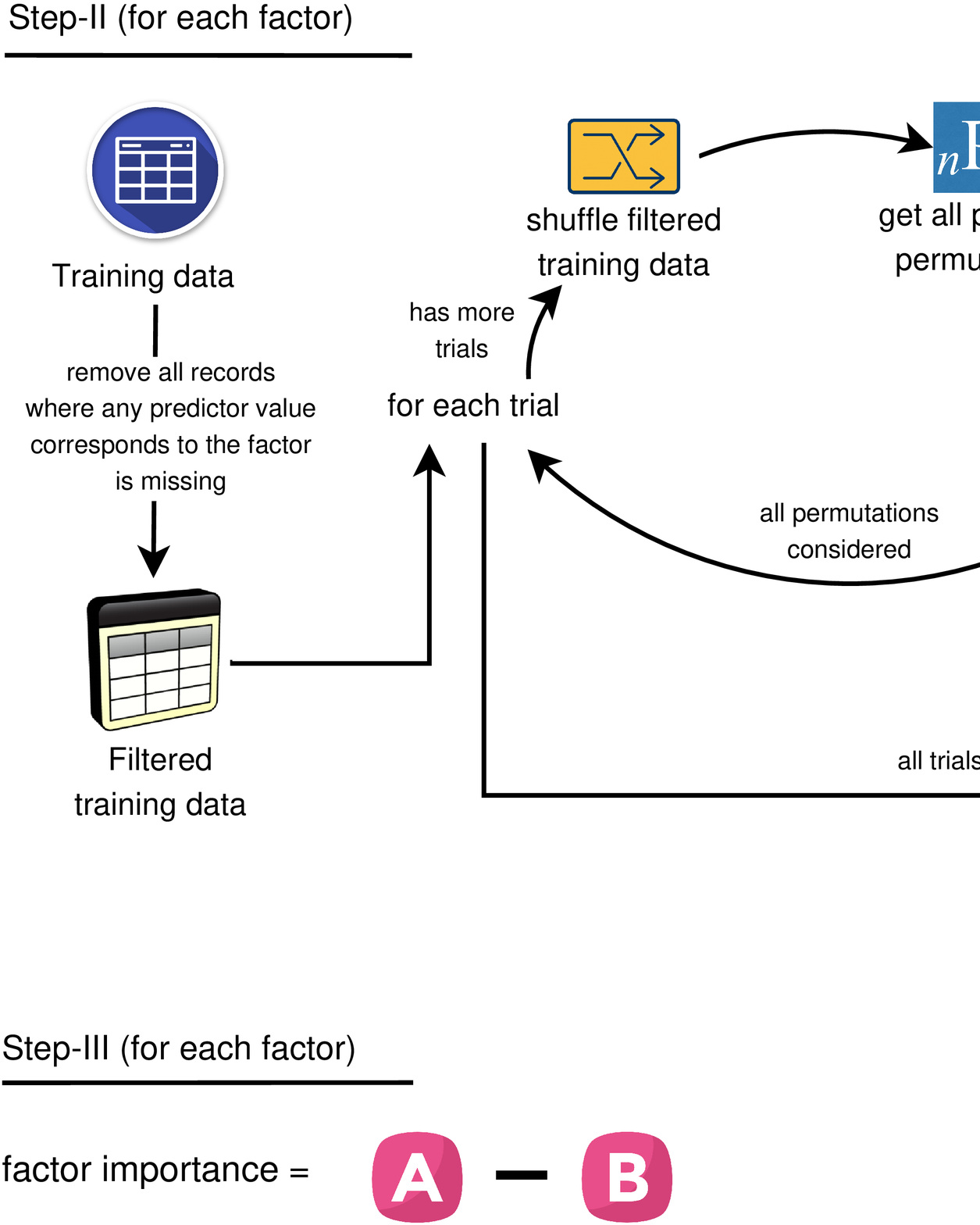}
	\caption{Methodology for measuring factor importance}
	\label{fig:methodology-factor-importance}
\end{figure}

\par
\Cref{fig:methodology-factor-importance} graphically explains the enhanced mechanism of measuring factor importance in the random wheel. In step-I, we have first measured the expected number of bins in the training dataset. Significantly, the number of bins is dependent on the order in which the training records are arranged. We have therefore randomly shuffled the training dataset multiple times and measured the average number of bins to obtain the expected bin count. Further division by the number of records in the training dataset provides the default bin ratio ($A$). In step-II, we have measured the expected bin count concerning each factor separately. It begins with filtering out all training records where missing values are observed in any attribute of the concerning factor. Let say the factor is $k$ and it is a combination of three attributes $\{a_1, a_2, a_3\}$. In order to measure the expected bin count for $k$, the filtration process removes all records where values are missing for either $a_1$ or $a_2$ or $a_3$. The next objective is to sort the filtered dataset to obtain the expected number of bins in the sorted dataset. Significantly, the sorting might produce a different number of bins based on the permutation of attributes considered for sorting. For example, the number of bins, while sorted based on $\{a_1, a_2, a_3\}$, would be different than that of in the sorted dataset based on $\{a_2, a_3, a_1\}$. Besides, the number of bins is dependent on the order of other attributes which are not considered in $k$ as well. We have therefore randomly shuffled the filtered dataset multiple times and measured the bin count for each attribute permutation separately. The average of all bin counts provides the expected number of bins concerning the factor $k$ here. We have finally divided the average bin count by the number of records in the filtered dataset to get the factor bin ratio ($B$). In step-III, the difference of default bin ratio ($A$) and factor bin ratio ($B$) provides the factor importance measure. It is significant to mention here that the factor importance calculated in this process may produce negative value as well. The random wheel discards these noisy factors as the number of bins increases in the sorted dataset in such cases.

\subsection*{Resultant Force}
The initial random wheel classifier \cite{khan2021randomwheel} combined multiple weighted elementary forces to obtain the resultant force to be applied on a wheel. Here we present the methodologies adopted in the enhanced version of this classifier to measure the resultant force.
\par

\begin{figure}[t]
	\centering
	\includegraphics[width=1.0\textwidth]{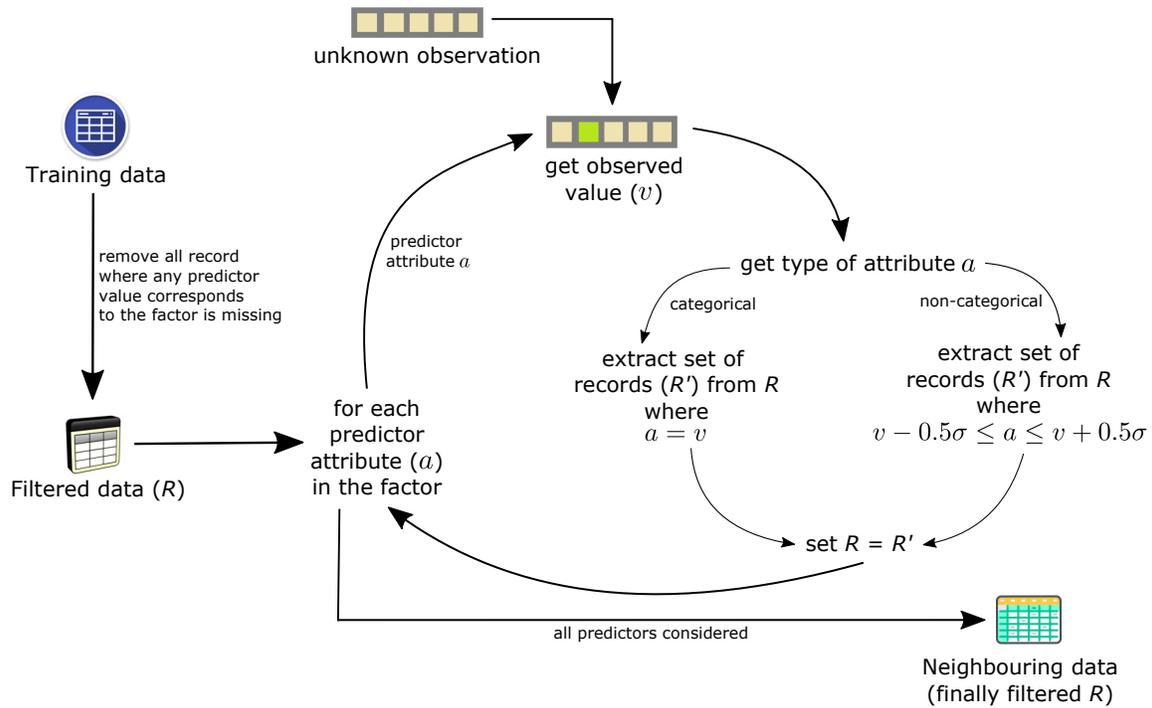}
	\caption{Methodology to obtain the neighbouring dataset}
	\label{fig:methodology-neighbouring-data}
\end{figure}

It is quite probable that an unknown observation would follow the same behaviour as its' neighbouring observations. The enhanced random wheel classifier, therefore, considers the neighbouring dataset to measure the weightage and elementary force with a mixed set of attributes. \Cref{fig:methodology-neighbouring-data} graphically presents an overview of the methods adopted for getting the neighbouring dataset here. Let us consider the same factor ($k$) again which is a combination of three attributes $\{a_1, a_2, a_3\}$ as mentioned earlier. We have initially removed all records in the training dataset where any attribute value corresponds to $a_1$, $a_2$ or $a_3$ is missing to obtain the filtered dataset. The next step thereafter extracts all records from the filtered dataset where the value of $a_1$, $a_2$ and $a_3$ are either same or within a predefined range. We have considered the record as a potential neighbour if the attribute value is the same in case of categorical attributes and within a range for non-categorical attributes. Let us say that the $a_1$ and $a_2$ are categorical and $a_3$ is non-categorical. The values of these attributes in unknown observation are say $v_1$, $v_2$ and $v_3$ respectively. We have considered all records of filtered dataset as neighbour in this case where the following three conditions are satisfied: (i) $value(a_1) = v_1$, (ii) $value(a_2) = v_2$, and (iii) $v_3 - 0.5\sigma \leq value(a_3) \leq v_3 + 0.5\sigma$. The $\sigma$ is the standard deviation of the non-categorical attribute $a_3$ here. This neighbouring dataset further helped us in measuring the factor weightage and magnitude of elementary force to be applied on a particular wheel of the trial.

\begin{figure}%
	\centering
	\subfloat[\centering Weightage \label{fig:fac-weightage}]{\includegraphics[width=0.29\textwidth]{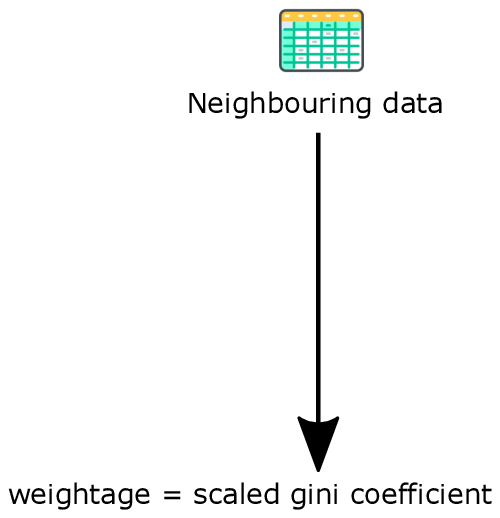} }%
	\subfloat[\centering Elementary force \label{fig:elem-force}]{\includegraphics[width=0.70\textwidth]{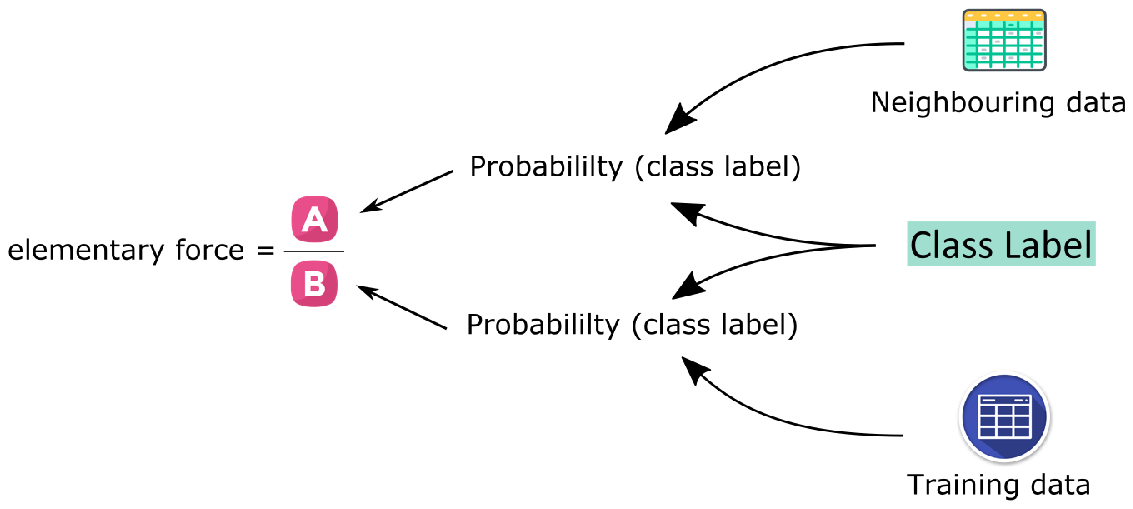} }%
	\caption{Methods for measuring weightage and elementary force of a factor using the neighbouring dataset}%
	\label{fig:weightage-and-elem-force}%
\end{figure}

\par
The random wheel provides more weightage to those factors that can induce more inequality into the dataset. As presented in \Cref{fig:fac-weightage}, we have calculated the scaled Gini coefficient observed in the neighbouring dataset to determine the factor weightage here. Say the neighbouring dataset corresponding to the factor $k$ contains $n_{1}$ and $n_{2}$ number of $+$ and $-$ class labels respectively. The $h^k$, as presented in \Cref{eq:scaled-gini}, would provide the weightage of factor $k$ in that case.

\begin{equation}
	h^k = 2 \times \dfrac{n_{1}^2 + n_{2}^2}{{({n_{1} + n_{2}})}^2} - 1
	\label{eq:scaled-gini}
\end{equation}

The elementary force ($f_j^k$) concerning to the factor $k$ indicates it's impact on class label $c_j$. The initial random wheel implementation \cite{khan2021randomwheel} considered the lift of association to measure this impact. We have followed the same approach here as well. However, the enhanced version of this classifier considers the neighbouring dataset along with the training dataset to measure the lift. \Cref{fig:elem-force} present an overview of this measurement. Let say the probability of class label $c_j$ in the neighbouring dataset and training dataset are $p_j^k$ and $\tilde{p}_j$ respectively. The magnitude of elementary force would therefore be as follows:

\begin{equation}
	f_j^k = \dfrac{p_j^k}{\tilde{p}_j}
	\label{eq:elem-force}
\end{equation}

Once all factor weightages and elementary forces concerning a particular class label are measured in a trial, the random wheel classifier combines them and applies the resultant force on the corresponding wheel. \Cref{eq:resultant-force} determines the resultant force to be applied on wheel $W_j$ representing class label $c_j$ in $i^{th}$ trial here.

\begin{equation}
	F_j^i = \sum_{\forall{k}}{h^k \times f_j^k}
	\label{eq:resultant-force}
\end{equation}

\subsection*{Explanation}
As mentioned earlier, the enhanced random wheel classifier considers the contribution of each attribute on the wheel rotation to explain the recommendation. \Cref{fig:methodology-explanation} presents an overview of the methods adapted here for this purpose. It starts with the creation of a list ($E$) that manages the contribution of all attributes for a recommendation. For each attribute, it iterates over the trials and aggregates the attribute contribution on the winner wheel in all trials.

\begin{figure}[t]
	\centering
	\includegraphics[width=1.0\textwidth]{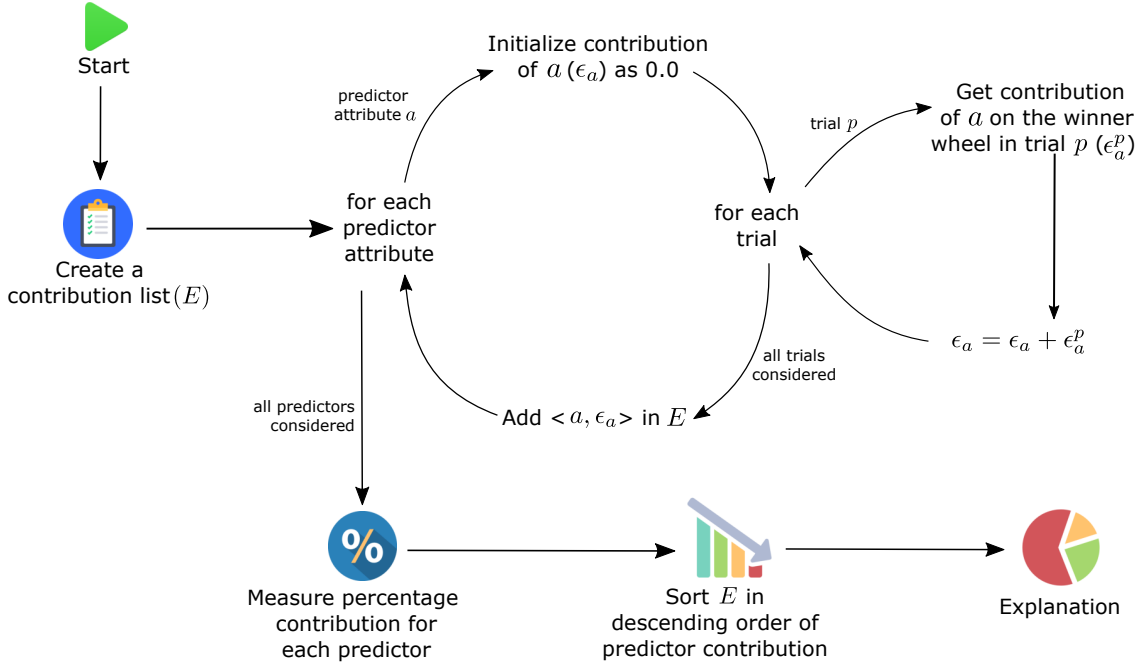}
	\caption{Methodology to obtain the explanation of recommendation}
	\label{fig:methodology-explanation}
\end{figure}

\par
Let us now present a case how the enhanced random wheel classifier calculates the contribution of a attribute, say $a_1$, in trial $p$. Say the trial $p$ considers four factors: \{\{$a_1$, $a_2$\}, \{$a_2$, $a_3$\}, \{$a_1$, $a_2$, $a_3$\}, \{$a_2$\}\}. In order to measure the contribution of $a_1$, it first filters all factors where $a_1$ exists. The classifier therefore selects \{\{$a_1$, $a_2$\}, \{$a_1$, $a_2$, $a_3$\}\} in this specific case. It then extracts the contribution of these two factors. Significantly, both weightage and elementary force are considered for measuring contribution here. Say, in case of the first factor (\{$a_1$, $a_2$\}), the weightage and the corresponding elementary force on the winner wheel are $h_{(a_1, a_2)}$ and $f_{(a_1, a_2)}$ respectively; whereas these are $h_{(a_1,a_2,a_3)}$ and $f_{(a_1,a_2,a_3)}$ for the second factor (\{$a_1$, $a_2$, $a_3$\}). \Cref{eq:contribution} here provides the contribution of attribute $a_1$ in trial $p$.

\begin{equation}
	\epsilon_{a_1}^p = \dfrac{h_{(a_1, a_2)} \times f_{(a_1, a_2)}}{2} + \dfrac{h_{(a_1,a_2,a_3)} \times f_{(a_1,a_2,a_3)}}{3}
	\label{eq:contribution}
\end{equation}

Here we consider that all attributes of a factor contribute uniformly. \Cref{eq:contribution} therefore divides the contribution of the first factor by two and second by three as there are two and three attributes available in them respectively. Once the attribute contributions in all trials are measured, the classifier finally calculates the percentage contribution for all of them and provide explanation in descending order of their contribution.

\printbibliography

\end{document}